\documentclass[article]{IEEEtran}

\usepackage[pdftex]{graphicx}
\ifCLASSINFOpdf
            \else
                  \fi

\usepackage{amssymb}
\usepackage[cmex10]{amsmath}
\usepackage{dsfont}
\def\y{\mathbf{y}}
\def\x{\mathbf{x}}
\def\w{\mathbf{w}}
\def\z{\mathbf{z}}
\def\xt{\mathbf{\tilde x}}

\usepackage{algorithmic}
\usepackage{algorithm}
\usepackage{amsthm}
\newtheorem{proposition}{Proposition}

\usepackage{multirow}

\usepackage[font=footnotesize]{subfig}

\usepackage[usenames,dvipsnames]{xcolor}

\newcommand{\remi}[1]{{\color{black} #1}}
\usepackage{url}

\begin{document}
\title{\remi{Non-convex Regularizations for Feature Selection in Ranking With Sparse SVM}}

\author{L\'ea~Laporte,
        R\'emi~Flamary,
        St\'ephane~Canu,
        S\'ebastien~D\'ejean,
        and~Josiane~Mothe,~\IEEEmembership{Fellows,~IEEE} \thanks{Manuscript received December 2012, revised June 2013, resubmitted July 2013.} \thanks{The authors would like to acknowledge the FR3424 Research Federation FREMIT and the Conseil G\'en\'eral of Midi-Pyr\'en\'ees (research project 10009108) for their financial support.}\thanks{L\'ea Laporte is with the Institut de Recherche en Informatique de Toulouse, UMR 5055, CNRS, Universit\'e de Toulouse, Toulouse 31062 Cedex 9, France e-mail: Lea.Laporte@irit.fr and with Nomao SA, Toulouse, 31500, France e-mail: lea@nomao.com.}\thanks{R\'emi Flamary is with the Laboratoire Lagrange, Universit\'e de Nice-Sophia Antipolis, Nice 06108, France (e-mail: remi.flamary@unice.fr)}\thanks{St\'ephane Canu is with INSA de Rouen, Saint-Etienne-du-Rouvray 76800, France (e-mail: stephane.canu@insa-rouen.fr)}\thanks{S\'ebastien D\'ejean is with the Institut de Math\'ematiques de Toulouse, UMR 5219, CNRS, Universit\'e de Toulouse, Toulouse 31062 Cedex 9, France (e-mail: sebastien.dejean@math.univ-toulouse.fr)}
\thanks{Josiane Mothe is with the Institut de Recherche en Informatique de Toulouse, UMR 5055, CNRS, Institut Universitaire de Formation de Ma\^itres, Universit\'e de Toulouse, Toulouse 31062 Cedex 9, France (e-mail: Josiane.Mothe@irit.fr) }
}

\markboth{Transactions on Neural Networks and Learning Systems,~Vol.~X, No.~X, December~2012}{Laporte \MakeLowercase{\textit{et al.}}: Non-convex Regularizations for Feature Selection in Ranking With SParse SVM}

\maketitle

\begin{abstract}
\remi{Feature selection in learning to rank has recently emerged as a crucial issue. Whereas several preprocessing approaches have been proposed, only a few works have been focused on integrating the feature selection into the learning process. In this work, we propose a general framework for feature selection in learning to rank  using SVM with a sparse regularization term. We investigate both classical convex regularizations such as $\ell_1$ or weighted $\ell_1$ and non-convex regularization terms such as log penalty, Minimax Concave Penalty (MCP) or $\ell_p$ pseudo norm with $p<1$. Two algorithms are proposed, first an accelerated proximal approach for solving the convex problems, second a reweighted $\ell_1$ scheme to address the non-convex regularizations. We conduct intensive experiments on nine datasets from Letor 3.0 and Letor 4.0 corpora. Numerical results show that the use of non-convex regularizations we propose leads to more sparsity in the resulting models while prediction performance is preserved. The number of features is decreased by up to a factor of six compared to the $\ell_1$ regularization. In addition, the software is publicly available on the web.\footnote{Software will be publicly available on the website of one of the author}}

\end{abstract}


\IEEEpeerreviewmaketitle

\section{Introduction}

\IEEEPARstart{L}{earning} to rank is a crucial issue in the field of information retrieval (IR). The main goal of learning to rank is to learn automatically ranking functions using a machine learning algorithm, in order to optimize the ranking of documents or web pages. Several algorithms have been proposed during the past decade \cite{Tieyanliu2011} that can combine a very large amount of features to learn ranking functions. 

Whereas the number of features that can be used by algorithms have increased, the issue of feature selection in learning to rank has emerged, for two main reasons.

First, as more and more features are incorporated into algorithms, not only the models become more difficult to understand, but also, they potentially have to deal with more and more noisy or irrelevant features. As feature selection is well-known in machine learning to deal with noisy and irrelevant features, it is seen as a quite natural way to solve this problem in learning to rank.

Second, the amount of training data used in learning to rank is substantial. As a consequence, learning a ranking function using algorithms is generally costly and can be time-consuming. Reducing the number of features, and thus the dimensionality of the problem, is a promising way to handle the issue of high computational cost.

Recent works have focused on the development of feature selection methods dedicated to learning to rank, which can be either preprocessing steps such as filter \cite{Geng:2007:FSR:1277741.1277811}\cite{Hua:2010:HFS:1772690.1772830}\cite{Yu:2009:EFW:1645953.1646100} and wrapper approaches \cite{Pan:2009:FSR:1645953.1646292}\cite{Yu:2009:EFW:1645953.1646100}\cite{dang2010}\cite{inpPaAiNaSa10a} or integrated to the learning algorithm, such as embedded approaches \cite{Sun:2009:RSR:1571941.1571987}\cite{10.1109/TC.2012.62}\remi{\cite{FSMRank}}.  In the latter case, the learning algorithm is called a sparse algorithm. In this paper, we consider an embedded approach for feature selection in learning to rank. 
\remi{We propose a general framework for feature selection in learning
  to rank using Support Vector Machines (SVM) and a regularization
  term to induce sparsity. We investigate both convex regularizaions such as $\ell_1$
  \cite{Tibshirani94regressionshrinkage} and non-convex
  regularizations such as MCP \cite{MCP}, log or $\ell_p$, $p<1$
  \cite{Kloft:2011:LPN:1953048.2021033}. To the best of our knowledge,
  this is the first work that investigates the use of non-convex
  penalties for feature selection in learning to rank. We first
  propose an accelerated Forward Backward Splitting algorithm in
  order to solve the $\ell_1$-regularized problem. Then, we propose a
  reweighted $\ell_1$ algorithm to handle the non-convex penalties
  that benefits from the first algorithm.  We conduct  intensive experiments on the Letor 3.0 and 4.0 corpora. Our convex algorithm leads to similar performance than the state-of-the-art methods. We show that the second algorithm that uses non-convex regularizations is a very competitive feature selection method, since it provides as good results as convex approaches but is much more performant in terms of sparsity. Indeed, it provides similar values of evaluation measures while using half as many features in average.}

This paper is organized as follows. Section 2 presents the state of
the art for learning to rank algorithms, feature selection methods,
sparse SVM and Forward Backward Splitting approaches. We formulate the
optimization problem in section 3. Section 4 introduces the algorithms
used to solve the optimization problems. We fully describe the datasets
used and the experimental protocol in section 5. In section 6, we
firstly analyze the ability of our approach to induce sparsity into
models. We secondly evaluate the performance of our framework in terms
of MAP and NDCG$@10$. We confront these results to those obtained with
two recent embedded feature selection methods.

\section{Related works}

Our work focuses on feature selection in learning to rank. We begin this section by presenting existing learning to rank algorithms. We provide an overview of feature selection methods dedicated to learning to rank and introduce feature selection using sparse regularized SVM.

\subsection{Learning to rank algorithms}

The learning to rank process consists in a training phase and a prediction phase.  In IR, the training data are composed of query-documents pairs represented by feature vectors. A relevance judgement between the query and the document is given as ground truth. The purpose of the training phase is to learn a model that provides the optimum ranking of documents according to their relevance to the query. The ability of the model to correctly rank documents for new queries is then evaluated during the prediction phase, on test data. Following is a short overview of learning to rank approaches and algorithms. A more complete introduction to learning to rank for IR can be found in the book by Liu \cite{Tieyanliu2011}. \\

Three approaches, called \textit{pointwise}, \textit{pairwise} and \textit{listwise}, have been proposed to solve the learning to rank problem. In the \textit{pointwise} approach, each instance is a vector of features  $x_i$  which represents a query-document pair. The ground truth can be either a relevance score $s \in \mathbb{R}$ or a class of relevance (such as "not relevant", "quite relevant", "highly relevant"). When dealing with a relevance score, learning to rank is seen as a regression problem. Some algorithms such as Subset Ranking \cite{Cossock:2006:SRU:2102571.2102627} have been proposed to solve it. When dealing with classes of relevance, learning to rank is considered as a classification problem or as an ordinal regression problem, depending on whether there is an ordinal relation between the classes of relevance. Some algorithms based on SVM \cite{Nallapati:2004:DMI:1008992.1009006} or on boosting \cite{conf/nips/LiBW07} deal with the classification problem. Crammer and Singer \cite{crammer:pranking} proposed an algorithm for ordinal regression. In the \textit{pairwise} approach, also referred as preference learning \cite{plbook}, each instance is a pair of feature vectors $(x_i,x_j)$ for a given query $q$. The ground truth is given as a preference $y \in \{-1,1\}$ between the two documents. For a given couple $(x_i,x_j)$, if $x_i$ is preferred to $x_j$, we note $x_i \succ_{q} x_j$ and then $y$ is set to $1$. In the contrary, if $x_j$ is preferred to $x_i$, we note $x_j \succ_{q} x_i$ and then $y$ is set to $-1$. It is thus a classification problem. Many algorithms have been developed to deal with this problem, such as RankNet \cite{Burges:2005:LRU:1102351.1102363} based on neural networks, RankBoost \cite{Freund:2003:EBA:945365.964285} based on boosting or RankSVM-Primal \cite{Chapelle:2010:EAR:1825381.1825398} and RankSVM-Struct \cite{Joachims:2002:OSE:775047.775067} based on SVM. Finally, the \textit{listwise} approach considers the whole ranked list of documents as the instance of 
the algorithm. Most works have focused on the proposal of new specific loss functions, based on the optimization of an IR metric or on permutations count in order to solve this kind of problem \cite{Cao:2007:LRP:1273496.1273513}\cite{Yue:2007:SVM:1277741.1277790}.\\

These approaches have been shown to be both efficient and effective to learn functions that ensure high ranking performance in terms of IR measures. Nevertheless, they may be suboptimal for use in real life with large scale data. Ranking functions deal with a very large amount of features, which raises three critical issues. First, as features may take time to compute, preprocessing steps such as the creation of training data may become time-consuming. Second, due to the high dimensionality of training data,  algorithms may not be scalable or they may take too much time for computation. Finally, there may be a significant amount of redundant or irrelevant features used by models, that can lead to suboptimal ranking performance. Thus, how to reduce the number of features to be used by algorithms has emerged as a crucial issue. Nevertheless, only few attempts have been made to solve this problem. In the following section, we propose an overview of existing feature selection methods in classification and 
learning to rank.

\subsection{Feature selection methods in learning to rank}

In classification, there are three kinds of feature selection methods called filter, wrapper and embedded. In filter methods, a subset of features is selected as a preprocessing step, independently of the predictor used for learning. In wrapper methods, the machine learning algorithm is used as a black box to score subsets of features according to their predictive power. The subset with the highest score is then chosen. Finally, in embedded methods, feature selection is performed within the training phase and incorporated to the algorithm. Embedded methods are generally specific to a given machine learning algorithm. A wide introduction to feature selection for classification is presented in the work of Guyon and Elisseeff \cite{Guyon:2003:IVF:944919.944968}. Feature selection methods for learning to rank have been developed in a similar way as in classification. We propose an overview of feature selection methods for learning to rank in the following section and we classify them into filter, wrapper and embedded categories in table \ref{FeatureSelectionMethods}.\\

To the best of our knowledge, the first proposal of a feature selection method dedicated to learning to rank is the work of Geng \MakeLowercase{\textit{et al.}} \cite{Geng:2007:FSR:1277741.1277811}. Their method is called Greedy search Algorithm for feature Selection (GAS) and belongs to filter approaches. For each feature, they first define its importance score: they rank instances according to feature values and evaluate the performance of the ranking list with a measure such as Mean Average Precision (MAP) or Normalized Discounted Cumulative Gain (NDCG). This evaluation measure is then used as the importance score for the feature. For each pair of features, they also define a similarity score, which is the value of the Kendall's $\tau$ between the rankings induced by the features of the pair. The Kendall's $\tau$ is defined as follow: if $x$ and $y$ are two features and $\mathcal{D}$ the number of documents pairs, then $\tau(x,y) = \frac{\#\{(d_s,d_t) \in \mathcal{D} | d_t \succ_x d_s \text{
and } d_t \succ_y d_s\}}{\#\{(d_s,d_t) \in \mathcal{D}\}}$ where $d_t \succ_x d_s$ indicates that the document $d_t$ is ranked above the document $d_s$ according to the value of feature $x$. An optimization problem is then formulated to select features by simultaneously maximizing the total importance score and minimizing the total similarity score. This optimization problem is solved by a greedy search algorithm. They show that the GAS algorithm can significantly improve the performance in terms of MAP or NDCG while reducing the number of features.

Hua  \MakeLowercase{\textit{et al.}} \cite{Hua:2010:HFS:1772690.1772830} later proposed a two-phase feature selection strategy. In a first step, they define the similarity between features in the same way as in the GAS algorithm. Features are then clustered into groups according to their similarity, by using a k-means approach. The number of clusters to be used is chosen according to a quality measure defined by the authors. In a second step, they propose to select a single representative feature from each cluster to learn the model. They use two delegation strategies for this purpose: a filter one  based on evaluation measure (BEM) and a wrapper one implied by the learning to rank method used (ILTR). The BEM delegation method selects the feature which ranking has the best evaluation score. The ILTR delegation method learns a linear model using a learning to rank algorithm. For each cluster, the representative feature is then the one with the highest weight in the ranking function. They show that BEM and ILTR 
techniques can significantly improve the performance in terms of NDCG@10 compared to models with no feature selection.\\

Some other works have focused on the development of wrapper approaches for feature selection on learning to rank. 

Pan  \MakeLowercase{\textit{et al.}} \cite{Pan:2009:FSR:1645953.1646292} proposed a method using boosted regression trees. In a similar way to \cite{Geng:2007:FSR:1277741.1277811}, they define an importance score for each feature and a similarity score for each pair of features. The importance score is the relative importance score as defined by Friedman \cite{Friedman00greedyfunction} for regression boosted trees. The similarity score is defined by the Kendall's $\tau$ between the vectors of values for the features of the pairs. The authors investigate three optimization problems: (1) to maximize the importance score, (2) to minimize the similarity score and (3) to simultaneously maximize the importance score and minimize the similarity score. These optimization problems are solved by a greedy approach. Experiments show that better results are obtained when only using the importance score than when using the importance and similarity scores. Moreover, they point out that a 30 features model achieves similar 
performance in terms of NDCG@5 than the complete model with 419 features. In a second approach, they propose a randomized feature selection with a feature-importance-based backward elimination. In practice, they create subsets of features, then iteratively train boosted trees and remove a percentage of features according to their NDCG@5 performance. The experimental results show that these methods achieve comparable performance than the complete model by using only 30 features.

Yu  \MakeLowercase{\textit{et al.}} \cite{Yu:2009:EFW:1645953.1646100} proposed two effective feature selection methods for ranking based on Relief algorithms \cite{Relief}. Relief algorithms are iterative methods that update the feature weights at each iteration, based on their importance.  The authors propose RankWrapper, a wrapper approach for training data with relative orderings, andRankFilter, a filter approach from training data with multi-level relevance classes. They also define new updating rules for the weights for each algorithm. Experiments on synthetic and benchmark datasets show that their method outperforms the GAS algorithm and can be used with large scale datasets.

Dang and Croft \cite{dang2010} proposed a feature selection technique based on the wrapper approach defined in \cite{Kohavi97wrappersfor}. They use a best-first search procedure to create subsets of features. For each subset, they train a model with a ranking algorithm. The output is defined as a new feature. A new feature vector is then created with the output of each subset and contains less features than the initial dataset. Models are trained using this vector with four well-known learning to rank algorithms: RankNet, RankBoost, AdaRank and Coordinate Ascent. Their experiments on Letor datasets show that they produce comparable performance in terms of NDCG@5 by using the smaller feature vector.

Finally, Pahikkala  \MakeLowercase{\textit{et al.}} \cite{inpPaAiNaSa10a} proposed an algorithm called greedy RankRLS, which is a wrapper approach based on the existing RankRLS algorithm. Subsets of features are created on which a leave-query-out cross-validation is performed by using the RankRLS algorithm. Results on the Letor 4.0 distribution show that the performance in terms of MAP and NDCG@10 are comparable to state-of-the-art algorithms with all the features.\\

Recently, embedded methods have been proposed to deal with the problem of feature selection. These approaches introduce a sparse regularization term in the formulation of the optimization problem.  Although sparse regularizations are widely used in classification to deal with feature selection, only a few attempts have been made to propose sparse-regularized learning to rank methods. Sun  \MakeLowercase{\textit{et al.}} \cite{Sun:2009:RSR:1571941.1571987} implemented a sparse algorithm called RSRank to directly optimize the NDCG. They propose a framework to reduce ranking to importance weighted pairwise classification. To achieve sparsity, they introduce a $\ell_1$-regularization term and solve the optimization problem using truncated gradient descent. Experiments on Ohsumed and TD2003 datasets show that only about a third of features remained after the selection. Moreover, the performance of the learned model is comparable or significantly better than the baselines, 
depending on the dataset and the measure used.

A more recent work of Lai  \MakeLowercase{\textit{et al.}} \cite{10.1109/TC.2012.62} proposed a primal-dual algorithm for learning to rank called FenchelRank. The authors formulate the sparse learning to rank problem as a SVM problem with a $\ell_1$-regularization term. They use the properties of the Fenchel Duality to solve the optimization problem. Basically, FenchelRank is an iterative algorithm that works in three steps. At each iteration, it first checks whether the stopping criterion is satisfied. If not, the algorithm then greedily chooses a feature to update according to its value. Finally, it updates the weights of the ranking model. Experiments were conducted on several datasets from the Letor 3.0 and Letor 4.0 collections. The authors show that FenchelRank leads to a good sparsity with sparsity ratios from 0.1875 to 0.5. It also provides comparable or significantly better results in terms of MAP and NDCG than state-of-art algorithms and RSRank. \remi{Finally, Lai \MakeLowercase{\textit{et al.}} \cite{FSMRank} recently proposed a new embedded algorithm for feature selection based on sparse SVM. This algorithm solves a joint convex optimization problem in order to learn ranking functions while automatically selecting the best features. They use a Nesterov approach to ensure fast convergence. They show that FSMRank can learn efficiently ranking models that outperform the GAS algorithm.}

 In classification, a large panel of embedded methods have been
 developed to learn sparse models with SVM. As far as we know,
 FenchelRank and FSMRank are the only ones to use sparse SVM for feature selection
 in learning to rank. Sparse SVM could widely and efficiently be
 adapted in this purpose. In this paper, we focuses on SVM methods
 with sparse-regularized term for which we propose a short overview in
 the following section.

\begin{table}
\renewcommand{\arraystretch}{1.3}
\caption{Classification of feature selection algorithms for learning to rank into filter, wrapper and embedded categories.}
 \label{FeatureSelectionMethods} 
 \centering
 \begin{tabular}{ccc}
 \hline
\textbf{Filter approaches} & \textbf{Wrapper approaches} & \textbf{Embedded approaches} \\
\hline\hline
GAS \cite{Geng:2007:FSR:1277741.1277811} & Hierarchical-ILTR \cite{Hua:2010:HFS:1772690.1772830} & RSRank \cite{Sun:2009:RSR:1571941.1571987} \\
Hierarchical-BEM \cite{Hua:2010:HFS:1772690.1772830} & BRTree \cite{Pan:2009:FSR:1645953.1646292} & FenchelRank \cite{10.1109/TC.2012.62} \\
RankFilter \cite{Yu:2009:EFW:1645953.1646100} & RankWrapper \cite{Yu:2009:EFW:1645953.1646100} & FSMRank \cite{FSMRank} \\
  & BFS-Wrapper \cite{dang2010} &\textit{[This work]}\\
  & GreedyRankRLS \cite{inpPaAiNaSa10a} & \\
 \hline
\end{tabular}
\end{table}

\remi{
\subsection{Learning as regularized empirical loss minimization}

The structural risk minimization is a useful and widely used induction principle in machine learning.
It states that the learning task can be defined
as the following  optimization problem (for a given positive constant $C$):
  \begin{equation}
    \label{eq:learningprob}
    \min_{\w,b} \quad C\sum_{i=1}^nL(\x_i^\top\w + b, y_i) + \Omega(\w)    .
  \end{equation}
The loss function $L(\cdot,\cdot)$ is
the data fitting term  measuring the discrepancy for all training
examples $\{\x_i,y_i\}$ between the predicted value
$f(\x)=\x^\top\w+b$ and the observed value $y$.
The second term
 $\Omega(\cdot)$ is a penalty providing regularization and controlling generalization ability through model complexity. }
Note that the training examples
may be also expressed as a matrix $X=[\x_1, \dots
,\x_n]^\top\in\mathbb{R}^{n\times d}$ and a vector
$\y=[y_1,\dots,y_n]^\top\in\mathbb{R}^n$ containing the output
objective values.

 Many choices are possible for the loss function $L(.,.)$ such as
  the
hinge loss, the squared hinge-loss or  the logistic loss for classification
problems
or  the $\ell_2$-loss or   the Hubert loss for a regression problem. 
 Similarly the regularization term $\Omega(\cdot)$ can be
the usual $\ell_2$-regularization also known
as ridge regularization ($\Omega_2(\w)=||\w||_2^2$ ) or the  $\ell_1$-regularization term also known as lasso
with $\Omega_1(\w)=||\w||_1=\sum_j |w_j|$. The latter has been proposed
in the context of linear regression by Tibishirani
\cite{Tibshirani94regressionshrinkage} in order to promote automated
feature selection. 

\remi{
\subsection{Feature selection via penalization}

When dealing with feature selection, the structural minimization principle still holds.
In this case, the relevant penalty from the statistical point of view is the $\ell_0$ regularization term.
It is defined as 
$\Omega_0(\w) =\|\w\|_0 =
\sum_{j=1}^d\mathds{1}_{\{w_{j=1}^d > 0\}}$ \emph{i.e.} the number of
non-zero components in vector $\w$.
But the minimization of such functional suffers severe drawbacks for optimization: it is not convex and not continuous nor differentiable (in zero).

A  way to tackle these issues is to relax the $\ell_0$ constraint
by replacing $\Omega_0(\w)$ by another penalty. 
A popular choice is the  $\ell_1$-regularization term.
It has several advantages such as being  convex and thus providing tractable optimization problems.

\remi{
However, if the $\ell_1$ lasso penalty alleviates the optimization issues raised
by the $\ell_0$ penalty, 
it brings some statistical concerns since it has been shown to be,  in certain cases,  inconsistent for variable selection and biased \cite{zou2006adaptive}.
A simple way to obtain nice statistical properties while preserving the computational efficiency of the $\ell_1$ minimization,
is to use a weighted lasso penalty by assigning different weights to each coefficient.}
This leads to the weighted $\ell_1$ regularization
$\Omega_\beta(\w)=\sum_{j=1}^d \beta_j|w_j| ,$ where the $\beta_j>0$ are the data-dependent
weights of each variable. Next step consists in proposing a suitable choice for these weights.

For regression problems, \cite{zou2006adaptive} proposed to use  $\beta_j=1/w_j^{LS}$ where $w_j^{LS}$ is the non-regularized least square estimator. 
This approach cannot be applied in our framework,  since the computation of the unpenalized solution is intractable.
Instead we propose to derive these weights from a non-convex relaxation the $\ell_0$
pseudo-norm
of the form
\begin{equation}
  \label{eq:regterm}
  \Omega(\w)=\sum_{j=1}^d g(|w_j|)
\end{equation}
where $g(\cdot)$ are non-convex functions.
Indeed, when it is well chosen, the non-convex nature of function $g$ will provide good statistical properties such as unbiasedness and oracle inequalities  \cite{MCP}, 
while the use of a  weighted $\ell_1$ implementation scheme will ensure nice computational behavior.
This can be obtained because the associated  problem
 have been shown to be part of a more general
optimization framework that can be solved using a simple reweighed
$\ell_1$ scheme with $\beta_j = g'(|w_j^*|)$, as presented in section \ref{sec:algor-nonc-regul},
where $g'$ denotes the  derivative of  $g$ 
and $w_j^*$ is a previously computed solution \cite{gasso2009recovering}. 

Among all possible choices for $g$,  we can cite the 
$\ell_p$  pseudo-norm with $p<1$   proposed by
\cite{fu1998penalized} and used more recently in compressed sensing
applications \cite{candes2008enhancing}. Another well known
approximation of the $\ell_0$ penalty is the log penalty that have been
introduced in \cite{Weston:2003:UZN:944919.944982} in the context of
variable selection. The Minimax  Concave Penalty (MCP) has been
proposed in \cite{MCP} in order to minimize the bias introduced by
classical $\ell_1$ regularizations. 
The smoothly clipped absolute deviation (SCAD)  is another popular choice.
These regularization terms are plotted in Figure \ref{fig:regterms} while their definition is recalled in the associated table.

\begin{figure}
  \centering
  \includegraphics[width=.8\columnwidth]{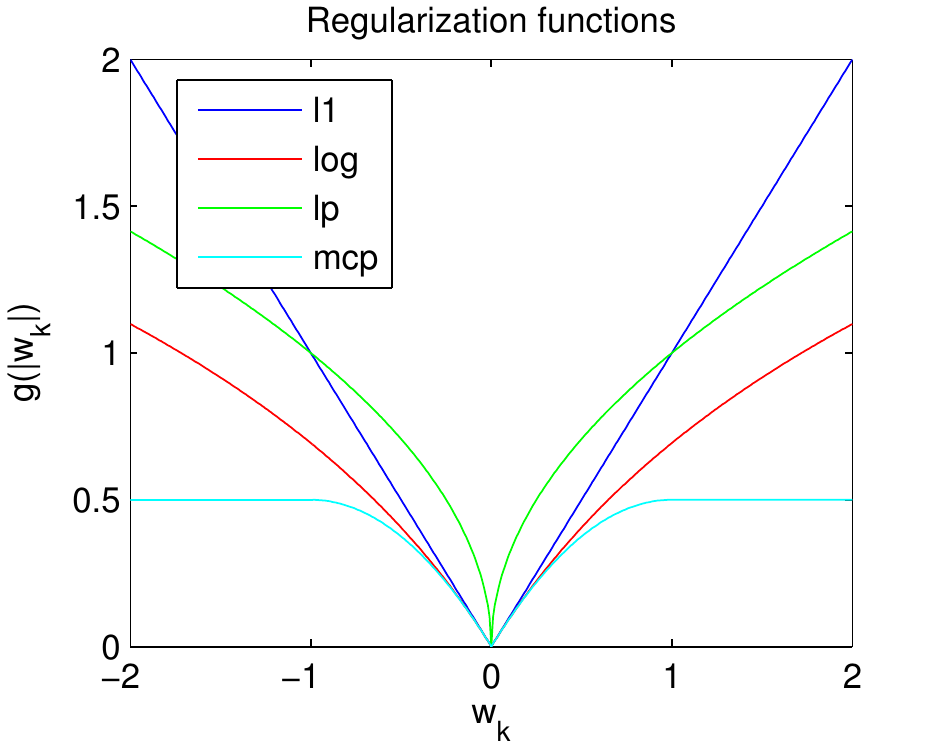}
  \begin{tabular}{|l|c|}
    \hline Reg. term & $g(|w_j|)$ \\\hline
     $ \ell_1$ & $ |w_j|$\\
  $\ell_p, p<1$& $|w_j|^p$\\
 log & $\log(\epsilon+|w_j|)$\\

 MCP & $\left\{\begin{array}{ll}\lambda |w_j|-|w_j|^2/2\gamma  & \mbox{ if } |w_j| \leq \gamma\lambda \\ \gamma\lambda^2/2 & \mbox{ if } |w_j| > \gamma\lambda \end{array}\right.$\\ \hline
   \end{tabular}

  \caption{Comparison of several  nonconvex regularization
    terms.  $\epsilon=1$ and $\gamma=1$
    are parameters respectively for the log and MCP regularizations.}
  \label{fig:regterms}
\end{figure}
}

In section 3, we formulate the sparse regularized problem. In section
4, we propose two algorithms to solve this non differentiable problem.

\section{Problem statement}
\subsection{Preference learning with SVM}

We consider a learning to rank problem in IR for which the documents ranking according to queries is to be optimized. Let $Q$ be the total number of queries and $D$ the total number of documents in the training dataset $T$. Then,  $\mathcal{Q} = \{q_k\}_{k=1,\hdots,Q}$ is the set of all queries and $\mathcal{D} = \{d_i\}_{i=1,\hdots,N}$ is the set of all documents. Each (query,document) pair is represented by a vector of features $ \phi(q_k,d_i) \in \mathbb{R}^d$ where $d$ is the number of features. Let $\w \in \mathbb{R}^d$ be the vector of weights of the learned model. We also define $r_k = \{(i,j)_{i,j=1,\hdots,N} | d_i \succ_{q_k} d_j \}$ as the subset of all the indices $(i,j)_{i,j=1,\hdots,N}$ for which there is a preference between $d_i$ and $d_j$ for the query $q_k$.

 The optimization problem to be solved in pairwise SVM for ranking is defined as in \cite{Joachims:2002:OSE:775047.775067}:
\begin{equation}
\label{Joachims}
\min_{\w \in \mathbb{R}^d,\xi_{i,j,k}} \frac{1}{2} \|\w\|_2^2 + C\sum_{i,j,k}\xi_{i,j,k} 
\end{equation}
under the constraints
\begin{displaymath}
\left \{
\begin{aligned}
\forall{(d_i,d_j)} \in r_1, \w^T(\phi(q_1,d_i) &- \phi(q_1,d_j)) > 1 - \xi_{i,j,1}\\
&\vdots \\
\forall{(d_i,d_j)} \in r_Q, \w^T(\phi(q_Q,d_i) &- \phi(q_Q,d_j)) > 1 - \xi_{i,j,Q}\\
\end{aligned}
\right.
\end{displaymath}
and 
\begin{displaymath}
\forall i, \forall j, \forall k, \xi_{i,j,k} \geq 0
\end{displaymath}

We can reduce this problem to a classification problem. Let  $\mathbb{I}^k = \{i^k\}_{i=1,\hdots,N \text{ and } ((i,.) \text{ or } (.,i)) \in r_k)}$ be the subset of indices of documents that take part of a preference relation for query $q_k$. We can then define the following vectors :
\begin{displaymath}
\textbf{D}= [ \{ i \in \mathbb{I}^1\} \hdots \{ i \in \mathbb{I}^Q\}]
\end{displaymath}
\begin{displaymath}
\textbf{Q} = [\underbrace{1\hdots1}_{card\{\mathbb{I}^1\}} \hdots \underbrace{k\hdots k}_{card\{\mathbb{I}^k\}} \hdots \underbrace{Q\hdots Q}_{card\{\mathbb{I}^Q\}}]
\end{displaymath}
where $\textbf{D} \in \mathbb{R}^{\sum_{k=1}^Q\text{card}\{\mathbb{I}^k\}} $ and $\textbf{Q} \in \mathbb{R}^{\sum_{k=1}^Q\text{card}\{\mathbb{I}^k\}} $.
Then the subset $\mathcal{P} \in \mathbb{R}^{n \times n}$ of all preference relations in the training dataset $T$ is defined as :
\begin{displaymath}
\mathcal{P} = \{ (s,t)_{s = 1,\hdots,n ; t = 1,\hdots,n} | (\textbf{D}(s),\textbf{D}(t)) \in r_{\textbf{Q}(s)=\textbf{Q}(t)}\}
\end{displaymath}
Each feature vector can be written as $\phi(.,.) = x_s$,$ s=1,\hdots,n$ and $X \in \mathbb{R}^{n \times d}$ is the matrix of all $x_s$ vectors.  The pairwise optimization problem is defined as:
\begin{equation}
\label{pairwise}
\min_{\w,\xi_p} \frac{1}{2}\|\w\|_2^2 + C\sum_{p=1}^P\xi_p
\end{equation}
under the constraints
\begin{displaymath}
\left \{
\begin{aligned}
&\xt_p \w \geq 1-\xi_p\\
&\xi_p \geq 0 \hspace{1in} \forall p = 1,\hdots,P \\
\end{aligned}
\right.
\end{displaymath}

where $\xt_p= x_s^T - x_t^T$ corresponds to a unique pair and
$\widetilde X =[\xt_1,\dots,\xt_1]^\top\in\mathbb{R}^{P\times d}$ is thus the matrix of all preferences pairs. Problem \ref{pairwise} is then equivalent to problem \ref{Joachims} and is written as a classification problem.
 By using the square hinge loss such as $\xi_p =
 \max(0,1-\xt_p^\top\w)^2$, the pairwise optimization problem finally is:
\begin{equation}
\min_{\w} \quad \frac{1}{2}\|\w\|_2^2 + C\sum_{p=1}^P\max(0,1-\xt_p^\top\w)^2
\label{pairwiseSVM}
\end{equation}
The use of the square hinge loss in this context as been proposed by Chapelle and Keerthi \cite{Chapelle:2010:EAR:1825381.1825398} for differentiability reasons when solving the pairwise problem in the primal.
\subsection{Sparse regularized SVM for preferences ranking}

To achieve feature selection in the context of SVM, a common solution
is to introduce a sparse regularization term. We propose in the
following to consider the Lasso formulation for feature selection,
that combine $\ell_1$-sparsity term and a square loss.

In classification, the Lasso SVM solves the following optimization problem:
\begin{equation}
\min_\w \quad\|\w\|_1+C\sum_{i=1}^n\max(0,1-y_i\x_i^\top \w)^2 
\label{lassoSVM}
\end{equation}

According to equations (\ref{pairwiseSVM}) and (\ref{lassoSVM}), we directly formulate the pairwise Lasso SVM by replacing the $\ell_2$-term by a $\ell_1$-term as:
\begin{equation}
\min_\w \quad\|\w\|_1+C\sum_{p=1}^P\max(0,1-\xt_i^\top \w)^2 
\label{RankSVM-Lasso}
\end{equation}

The optimization problem for pairwise learning to rank with Lasso SVM
is thus reduced to a Lasso classification problem on the matrix of
preferences. One critical issue may arise when using this
formulation. Indeed, as the $\ell_1$-norm is not
differentiable, this problem might be quite difficult to
solve. However, a large panel of methods and algorithms have been proposed in classification in order to solve it. Thus, we
argue that considering pairwise sparse SVMs  is perfectly well suited
to select features in learning to rank, for two reasons. 

Firstly, contrary to several proposed approaches such as GAS, sparse regularized SVM methods do not require extra developments of similarity and importance measures dedicated to learning to rank. Indeed, the feature selection is only based on the properties of the regularization term, so no additional assumptions are needed. Secondly, as they follow the SVM framework for classification, methods and algorithms used in classification can easily be adapted and applied to learning to rank with a few implementation efforts. In this paper, we propose to use  an adaptation of a Fast Iterative Shrinkage Tresholding Algorithm in order to solve the sparse regularized optimization problem and to proceed to feature selection. We present this algorithm in the following section. We also propose to use non-convex regularization instead of $\ell_1$-penalty in order to counter the statistical issues that may arise. We propose a second algorithms in the following section to deal with the non-convex regularizations.

\section{Learning
  preferences with sparse  SVM}

\remi{In this section, we discuss the proposed methods for learning
  preferences with sparse SVM.  Firstly, we introduce Forward-Backward Splitting
algorithm, that is a well known approach for solving the non-differentiable weighted-$\ell_1$ regularized problem. Secondly, this algorithm is
adapted to the problem of preference learning and its convergence is
proved. Finally we propose a general approach for solving the learning
problem with non-convex regularization terms.  }

\subsection{Forward-Backward Splitting  Algorithms for feature
  selection}

Forward-Backward Splitting (FBS) algorithms have been proposed initially to solve
non-differentiable optimization problems such as $\ell_1$-norm
regularized learning problems. A good introduction to this kind of
algorithm is given in \cite{bach2011convex}. When minimizing a problem
of the form:
\begin{equation}
   \label{eq:probAG}
  \min_{\w\in\mathbb{R}^d}\quad J_1(\w)+\lambda\Omega(\w) 
\end{equation}
where $J_1(\cdot)$ is a differentiable objective function with a
Lipschitz continuous gradient and $\Omega(\cdot)$ is a convex
regularization term having a closed form proximity operator, the proximity operator of regularization $\mu\Omega(\cdot)$ is
defined as
\begin{equation}
  \label{eq:proximaloperator}
  \text{Prox}_{\mu\Omega}(\mathbf{z})=\arg\min_\w \quad
  \frac{1}{2}\|\mathbf{z}-\w\|_2^2+\mu\Omega(\w)
\end{equation}
FBS algorithms are iterative methods that compute at each iteration the
proximity operator of the regularization term on a
gradient descent step with respect to the differentiable function, thus
leading to the following update
\begin{equation}
\w^{k+1}=\text{Prox}_{\frac{\lambda}{L}\Omega}\left(\w^k-\frac{1}{L}\nabla
J_1(\w^k)\right)\label{eq:step}
\end{equation}
where $\frac{1}{L}$ is a gradient step and $L$ has to be a  Lipschitz
constant of $\nabla J_1$ in order to ensure convergence. Note that one
can easily compute the proximity operator of the $\ell_1$-regularization that is of the following form:
\begin{equation}
  \label{eq:proximal_l1}
  \begin{aligned}
  \text{Prox}_{\lambda\|\cdot\|_1}(\w)_j&=\max\left(0,1-\frac{\lambda}{|w_j|}\right)w_j\\
  &=\text{sign}(w_j)(|w_j|-\lambda)_+\quad
  \forall j\in 1,\dots,d
  \end{aligned}
\end{equation}
\remi{The weighted $\ell_1$ regularization has a similar proximity operator
\begin{equation}
  \label{eq:proximal_l1}
  \begin{aligned}
  \text{Prox}_{\lambda\Omega_\beta}(\w)_j&=\max\left(0,1-\frac{\lambda\beta_j}{|w_j|}\right)w_j\\
  &=\text{sign}(w_j)(|w_j|-\lambda\beta_j)_+\quad
  \forall j\in 1,\dots,d
  \end{aligned}
\end{equation}}
This
algorithm, also known as Iterative Shrinkage Thresholding
Algorithms (ISTA), has been proposed  to solve linear inverse problems with
$\ell_1$-regularization as presented in
\cite{Beck:2009:FIS:1658360.1658364}.  In their paper, Bech and Teboulle also
address one limitation of this kind of approach: speed of
convergence. 
Although
 these algorithms are able to deal with large-scale data, they may
 converge slowly. They proposed to use a multistep version of the
 algorithm called Fast Iterative Shrinkage Thresholding
Algorithms (FISTA) that will converge more quickly to the optimal objective
value. This algorithm can be seen in algorithm \ref{alg:afbs}.
\begin{algorithm}[t]
  \begin{algorithmic}[1]
    \STATE Initialize $\w^0$ \STATE Initialize $L$ as a Lipschitz
    constant of $\nabla J_1(\cdot)$ \STATE $k=1$, $\z^1=\w^0$ ,
    $t^1=1$ \REPEAT \STATE $\w^k\leftarrow
    \text{Prox}_{\frac{\lambda}{L}\Omega}(\z^k-\frac{1}{L}\nabla
    J_1(\z^k))$ \STATE $t^{k+1}\leftarrow
    \frac{1+\sqrt{1+4(t^{k})^2}}{2}$ \STATE $\z^{k+1}\leftarrow \w^k+
    \left(\frac{t^k-1}{t^{k+1}}\right)(\w^k-\w^{k-1})$ \STATE
    $k\leftarrow k+1$ \UNTIL{Convergence}
  \end{algorithmic}
\caption{Accelerated FBS algorithm}
\label{alg:afbs}
\end{algorithm}

\subsection{FBS for sparse preference learning}
In this section, we discuss how we adapted the FISTA algorithm  to the
problem of
preference learning with \remi{$\ell_1$ and weighted $\ell_1$-regularized} SVM. 

First we note that
problem (\ref{RankSVM-Lasso}) is a sum of a differential function, the
data fitting loss and a non-differentiable $\ell_1$-regularization.
We then solve the equivalent problem as in (\ref{eq:probAG}) with
$\Omega(\w)=||\w||_1$, $\lambda=1/C$ and
$J_1(\w)=\sum_{p=1}^P\max(0,1-\xt_p^\top\w)^2$.

In order to ensure convergence of the algorithm, the cost
function $J_1(\w)$ must have a Lipschitz continuous gradient. Then, we just have to prove proposition \ref{prop:lipschitz}: 
\begin{proposition}
\label{prop:lipschitz}
Let $J_1(\w)$ the square Hinge loss
\[
J_1(\w) = \sum_{p=1}^P\max(0,1-\xt_p^\top\w)^2
\]

Then its gradient 
\[
 \nabla J_1(\w)= -2\sum_{p=1}^P\xt_p\max(0,1-\xt_p^\top\w)
\]
 is Lipschitz and continuous.
\end{proposition}

\begin{IEEEproof}
The squared Hinge loss is gradient Lipschitz if there exists a
constant $L$ such that:
\begin{displaymath}
\|\nabla J_1(\w_1) - \nabla J_1(\w_2)\|_2 \leq L \|\w_1 - \w_2\|_2 
\quad \forall \w_1,\w_2 \in d.
\end{displaymath}
The proof essentially relies on showing that $\xt_i\max(0,1-\xt_i^\top \w)$
is Lipschitz itself \emph{i.e} there exists $L^\prime \in \mathbb{R}$ such that
\begin{align}\nonumber
&\|\xt_i \max(0,1-\xt_i^\top \w_1) - \xt_i \max(0,1-\xt_i^\top \w_2)  \| \\ &
\leq L^\prime \|\w_1 - \w_2\| \nonumber
\end{align}
Now let us consider different situations. For a given 
$\w_1$ and $\w_2$, if 
  $1-\xt_i^T\w_1 \leq 0$
and   $1-\xt_i^T\w_2 \leq 0$, then the left hand side is equal to $0$
and any $L^\prime$ would satisfy the inequality.
 If 
  $1-\xt_i^T\w_1 \leq 0$
and   $1-\xt_i^T\w_2 \geq 0$, then the left hand side (lhs) is
\begin{eqnarray}
  \label{eq:lipschitz}\nonumber
 lhs &=& \|\xt_i\|_2 (1-\xt_i^\top \w_2) \\ \nonumber
&\leq& \|\xt_i\|_2(\xt_i^\top\w_1 -\xt_i^\top \w_2)\\ \nonumber
&\leq& \|\xt_i\|_2^2 \|\w_1-\w_2\|_2 \nonumber
\end{eqnarray}
A similar reasoning yields to the same bound when 
 $1-\xt_i^T\w_1 \geq 0$ and $1-\xt_i^T\w_1 \leq 0$
and   $1-\xt_i^T\w_2 \geq 0$
and   $1-\xt_i^T\w_2 \geq 0$. Thus,  $\xt_i\max(0,1- \xt_i^\top \w)$ is Lipschitz
with a constant $\|\xt_i\|^2$. Now, we can conclude
the proof by stating that $\nabla_\w J$ is Lipschitz
as it is a sum of Lipschitz function and
the related constant is $\sum_{i=1}^n\|\xt_i\|_2^2$.  
\end{IEEEproof}

We thus have proved than $J_1(\w)$ has a Lipschitz continuous gradient, which ensure the convergence of the algorithm. The gradient of $J_1(\w)$ is easy to compute and can be used as it in the FISTA algorithm. We thus can use the FISTA algorithm to solve the $\ell_1$ and weighted-$\ell_1$-regularized SVM problems. We called this algorithm RankSVM-$\ell_1$.

\remi{

\subsection{Algorithm for non-convex regularization}
\label{sec:algor-nonc-regul}

When using a non-convex regularization term as presented in equation
\eqref{eq:regterm}, the previous algorithm
cannot be used. We propose in this case to adapt a general purpose
framework that has been proposed in \cite{gasso2009recovering}. The
main idea behind this framework is to cast the regularization term as
a difference of two convex
functions. Convergence to a stationary point has been proven on this
particular class of problems when performing a primal/dual
optimization \cite{tao1998dc}.

The approach introduced in \cite{gasso2009recovering} can also be seen
as a Majorization Minimization method
\cite{hunter2004tutorial}. Indeed, one can clearly see in Figure
\ref{fig:regterms} that all of the proposed regularization terms are
concave in their positive orthant.  This implies that for a fixed
point $u_0>0$
\begin{equation*}
\forall u>0,\quad\quad g(u) \leq g(u_0) + g'(u_0)(u-u_0)
\end{equation*}
The algorithm consists in minimizing iteratively the majoration of the
cost function. When removing the constant term the optimization
problem for  iteration $k+1$ is
\begin{equation}
  \label{eq:probmajo}
 \w^{k+1}=   \arg\min_{\w\in\mathbb{R}^d}\quad J_1(\w)+\lambda\sum_i \beta_j|w_j|
\end{equation}
where $\beta_i=g'(|w_i^k|)$ is computed using the solution at the previous
iteration. This approach is extremely interesting in our case as we
can readily use the efficient algorithm proposed for the weighted
$\ell_1$ regularization. Moreover, one can use a warm-starting scheme
for initializing the solver at the previous iteration. The resulting algorithm
is given in algorithm \ref{alg:ncopt} and the derivative functions
$g'(\cdot)$ for
the non-convex regularizations  are given in table \ref{tab:deriv}.

\begin{algorithm}[t]
  \begin{algorithmic}[1]
    \STATE Initialize $\w^0$ and $k=1$ 
\STATE Initialize $\beta_j=1, \forall_j$
\REPEAT 

\STATE $\w^k\leftarrow$ Minimize majorization \eqref{eq:probmajo} using algorithm \ref{alg:afbs}.
\STATE $\beta_j\leftarrow g'(|w_j^k|), \forall j$
\STATE $k\leftarrow k+1$
 \UNTIL{Convergence}
  \end{algorithmic}
\caption{Solver for non-convex regularization}
\label{alg:ncopt}
\end{algorithm}

\begin{table}
  \centering
  \caption{Derivatives of the nonconvex regularization terms}
\label{tab:deriv}
  \begin{tabular}{|l|c|}
        \hline Reg. term & $g'(|w_j|)$ \\\hline
 $\ell_p, p<1$& $p|w_j|^{p-1}$\\
 log & $1/(\epsilon+|w_j|)$\\
 MCP & $max(1-|w_j|/(\gamma\lambda),0)$ \\\hline
  \end{tabular}

\end{table}

}

\section{Experimental framework}

A set of numerical experiments have been conducted on benchmark datasets to evaluate the performance of the framework we proposed. In this section, we provide a full description of the datasets and the measures used. We also present the experimental protocol.

\subsection{Datasets}

We conduct our experiments on Letor 3.0 and Letor 4.0 collections. These are benchmarks dedicated to learning to rank. Letor 3.0 contains seven datasets: Ohsumed, TD2003, TD2004, HP2003, HP2004, NP2003 and NP2004. Letor 4.0 contains two datasets, MQ2007 and MQ2008. Their characteristics are summarized in table \ref{DistDescription}. Each dataset is divided into five folds, in order to perform cross validation. For each fold, we dispose of train, test and validation sets.

\begin{table*}[!t]
\caption{\label{DistDescription} Characteristics of Letor 3.0 and Letor 4.0 distributions.} 
\centering
 \begin{tabular}{c | c  c  c  c | c  c }
 \hline
\multirow{2}{*}{\textbf{Dataset}}& \multicolumn{4}{c|}{\textbf{Number of}} & \multicolumn{2}{c}{\textbf{Average number (per query) of}}  \\
 & \textbf{features} & \textbf{queries} & \textbf{query-document pairs} & \textbf{preferences} & \textbf{documents} & \textbf{relevant documents} \\
\hline\hline
TD2003 & 64 & 50 & 49058 & 398562 & 981.1 & 8.1\\
NP2003 & 64 & 150 & 148657 & 150495 & 991 & 1\\
HP2003 & 64 & 150 & 147606 & 181302 & 984 & 1\\
TD2004 & 64  & 75 & 74146 & 1079810 & 988.6 & 14.9\\
NP2004 & 64 & 75 & 73834 & 75747 & 992.1 & 1.1\\
HP2004 & 64 & 75 & 74409 & 80306 & 984.5 & 1\\
Ohsumed & 45 & 106 & 16140 & 582588 & 152.2 & 45.6\\
MQ2007 & 46 & 1692 & 69623 & 404467 & 41.1 & 10.6\\
MQ2008 & 46 & 784 & 15211 & 80925 & 19.4 & 3.7\\
\hline
\end{tabular}
\end{table*}

\subsection{Evaluation measures}
We evaluate the ranking performance of our approach using the Mean Average Precision (MAP) and the Normalized Discounted Cumulative Gain (NDCG).
MAP is a standard evaluation measure in information retrieval that works with binary relevance judgements: relevant or not relevant. It is based on the computation of the precision at the position $k$ which represents the fraction of relevant documents at the position $k$ in the ranking list for a query $q$:
\begin{displaymath}
P_q@k = \frac{\#\text{relevant documents within the k top documents}}{k}
\end{displaymath}

The Average Precision (AP) at the position $k$ is then defined for the query $q$ as:
\begin{displaymath}
AP_q= \frac{\sum_{i=1}^{k}P@i.\mathds{1}_{\{\text{document i is relevant}\}}}{\# \text{relevant documents for the query q}}
\end{displaymath}

The MAP is defined as the average of AP over all the queries:
\begin{displaymath}
MAP = \frac{1}{Q}\sum_{q=1}^Q AP_q
\end{displaymath}

Unlike MAP, the NDCG can deal with more than two levels of relevance.  Let $r(i)$ be the relevance level of the document at position $i$. Given a query $q$, the Discounted Cumulative Gain at position $k$ is defined as:
\begin{displaymath}
DCG_q@k = \sum_{i=1}^{k}\frac{2^{r(i)}-1}{log_2(i+1)}
\end{displaymath}
DCG can take values greater than 1. A normalization term is then introduced to set values from 0 to 1:
\begin{displaymath}
NDCG_q@k = \frac{1}{Z_k}DCG_q@k
\end{displaymath}
where $Z_k$ is the maximum value of DCG@k.

We also evaluate the ability of our approach to promote sparsity. To this purpose, we compute the sparsity ratio, which is the fraction of remaining features in the model after selection. For each fold $f \in {1,\dots,N_\mathcal{T}}$ of the dataset $\mathcal{T}$, $N_\mathcal{T}$ the number of folds, we define the sparsity ratio $SR_f$ as:
\begin{displaymath}
SR_f = \frac{\#\text{remaining features in the learned model}}{\#\text{features of the given dataset}}
\end{displaymath}
We do not consider features that are zero for all the queries. Thus, the total number of features of a given dataset can be smaller that indicated in table \ref{DistDescription}. The sparsity ratio of the algorithm $\mathcal{A}$ for a given dataset $\mathcal{T}$ is the average of SR over all the folds:
\begin{displaymath}
SR_{(\mathcal{A},\mathcal{T})} = \frac{1}{N_\mathcal{T}}\sum_{f = 1}^{N_\mathcal{T}} SR_f 
\end{displaymath}

\subsection{Experimental protocol}
For each dataset, we first train the algorithms on the training set
with different values of $C$ on a grid. For each fold, the $C$ value
that leads to the best MAP performance on the validation set is
chosen. The model trained with this $C$ value is used for prediction
on the test set. We compute the MAP and the NDCG$@10$ on the test
dataset. \remi{We then compare the convex algorithm and the
  non-convex algorithm to the state-of-the-art methods FenchelRank
  and FSMRank. We do not compare our method to the GAS algorithm, since it has been proven to be outperformed by the FSMRank algorithm \cite{FSMRank}. We run the Windows/MsDos FenchelRank
  executable provided on the author's personal web
  page\footnote{scholat.com/\textasciitilde hanjiang. Last visited on
    12/09/2012} and the matlab code of FSMRank provided by the authors
  on demand. We use the same grid as the authors to tune the
  parameters. Please note that since we use the MAP instead of the
  NDCG$@10$ to choose the optimal value of $r$ on the validation, we
  obtain different models and results than in
  \cite{10.1109/TC.2012.62} and \cite{FSMRank}. Finally, we set $\gamma=2$ for the MCP
  penalty and $\epsilon=0.1$ for the log penalty, that are values commonly used  in the community.}

For each experiment, we use the paired one-sided Student test in order to evaluate the significance of our results. A result is significantly better than another if the p-value provided by the Student test is lower than 5\%. Results on performance in terms of sparisty ratio are illustrated by a spider (or radar) plot. Spider plots allow us to easily compare the behavior of several algorithms on several datasets according to a given measure. Each branch of the plot represents a dataset while each line stands for an algorithm.

\section{Results and discussion}

\remi{In this section, we compare our convex and non-convex frameworks to state-of-the-art methods. Firstly, we analyze the performance of the non-convex framework in terms of sparsity ratio. Scondly, we show that using non-convex regularizations leads to similar results both in terms of MAP and NDCG$@10$. Finally, we confront the sparsity ratio and the performance in terms of IR measures to demonstrate that non-convex regularizations are truly competitive compared to state-of-the-art approaches.

\subsection{Sparsity ratio}

As we stated in the introduction, feature selection is a key issue in learning to rank. We aim at providing effective methods that can learn high quality models while automatically selecting a few number of highly informative features.
The main goal of using non-convex regularizations is to sharply reduce the amount of features used in ranking models. In this section, we analyze the sparsity ratio we obtain by using non-convex penalties and we compare them with the two algorithms FenchelRank and FSMRank and our $\ell_1$ algorithm.

Table \ref{SparsityRatio} presents the sparsity ratio obtained with FenchelRank, FSMRank, the $\ell_1$regularization and the three non-convex penalties log, MCP and $\ell_{0.5}$. We restrain our analysis to this value of $p$ for readability reasons. Figure \ref{fig:removeSparsity} presents the spider plot of $1 - $ sparsity ratio, that is the ratio of removed features. The larger this measure is, the better the algorithm is in  order to induce sparsity into models.

We firstly observe on both table \ref{SparsityRatio} and figure
\ref{fig:removeSparsity} that, in average, methods that use convex
penalty are not as sparse as those using non-convex
regularizations. Methods that use the $\ell_1$ penalty are the less
sparse. In particular, FSMRank leads to higher sparsity ratio on most
of the datasets, which means that the learned models contain much more
features than those learned by the other methods. The MCP penalty
appears to be the less sparse of the non-convex penalties. This is not
really surprising since the MCP penalty has been initially proposed,
not as a feature selection approach, but as a way to minimize the bias
induced by $\ell_1$ regularization.

When considering the average sparsity ratio, the use of log and $\ell_p$ penalties makes sense. These two non-convex penalties lead to the smallest sparsity ratio. The learned models select in average half of the features used by convex regularizations on all datasets. When considering each dataset independently, log and $\ell_p$ penalties select up to twelve times less features than the convex ones. These penalties are then truly performant methods to achieve feature selection. The log penalty is particularly effective for inducing sparsity on HP2003, TD2003 and the MQ datasets. For these later datasets, it selects from around six to twelve times fewer features than the state-of-the-art algorithms. The log penalty is the most effective for inducing sparsity on Ohsumed, HP2004 and NP2004 datasets. It can frequently selects from quarter ot half less features than the convex regularizations. More precisely:
\begin{itemize}
 \item on Ohsumed, the log penalty selects from half to third as many features than convex and MCP penalties,
 \item on MQ2008, there were four to six times fewer features used by $\ell_{0.5}$ than by convex or MCP regularizations, while the log penalty selects two to three times fewer features,
 \item on MQ2007, the log penalty selects half as many features as convex and MCP penalties while $\ell_{0.5}$ selects from ten to twelve times fewer features that convex regularizations and MCP,
 \item on HP2004, the two non-convex penalties use from a quarter to an half as many features as MCP and convex regularizations,
 \item on NP2004, the log penalty selects from an half to a third as many features as MCP and convex regularizations,
 \item on TD2004 and HP2003, there were two to three times fewer features used by  $\ell_{0.5}$ than by MCP and convex regularizations,
 \item on TD2003, the non-convex $\ell_{0.5}$ penalty use half as many features as MCP and convex regularizations.
\end{itemize}

Non-convex penalties are then shown to be very competitive methods when considering the number of selected features. The difference of sparsity ratio observed between datasets is due to the intrinsec difference between datasets. Ohsumed dataset, Letor 4.0 and Letor 3.0 collections do not all use the same features. Althought features are similar for HP, NP and TD datasets, those are not related to similar retrieval tasks. The amount of relevant features may vary from a kind of datasets to an other, so does the relevant features themselves. Nevertheless, we may reasonbly expect to select same features on datasets related to the same tasks. The different performance in terms of sparsity ratio of a given algorithm among the datasets should not be seen as drawback of the method, but as the specificity of the dataset.

Removing a large amount of features may not be accurate if it leads to a degradation of prediction quality. In the following section, we compare the performance in terms of IR measures between non-convex and convex regularizations.

\begin{table}
\caption{Comparison of sparsity ratio between convex and non-convex regularizations and state-of-the-art algorithms.}
\label{SparsityRatio}
 \begin{center}
  \begin{tabular}{c |c c |c c c c}
   \hline
   \textbf{Dataset} &  \textbf{Fenchel} & \textbf{FSM} &\textbf{$\ell_1$} & \textbf{MCP}& \textbf{log}  &\textbf{$\ell_{0.5}$}\\
   \hline\hline
Ohsumed&0.23&0.41&0.28&0.34&\textbf{0.12}&0.22\\
MQ2008&0.3&0.42&0.39&0.27&0.09&\textbf{0.07}\\
MQ2007&0.58&0.64&0.6&0.19&0.29&\textbf{0.05}\\
HP2004&0.19&0.26&0.17&0.24&\textbf{0.06}&0.12\\
NP2004&0.27&0.37&0.43&0.32&\textbf{0.13}&0.23\\
TD2004&0.46&0.67&0.42&\textbf{0.28}&0.36&0.3\\
HP2003&0.27&0.48&0.25&0.39&0.27&\textbf{0.16}\\
NP2003&0.23&0.44&0.48&0.36&\textbf{0.23}&0.27\\
TD2003&0.53&0.76&0.48&0.40&0.34&\textbf{0.20}\\
\hline
\textit{Mean}&\textit{0.34}&\textit{0.49}&\textit{0.39}&\textit{0.31}&\textit{0.21}&\textit{\textbf{0.18}}\\
   \hline
  \end{tabular}
 \end{center}
\end{table}

\begin{figure}[!t]
 \centering
 \includegraphics[width=0.5\textwidth]{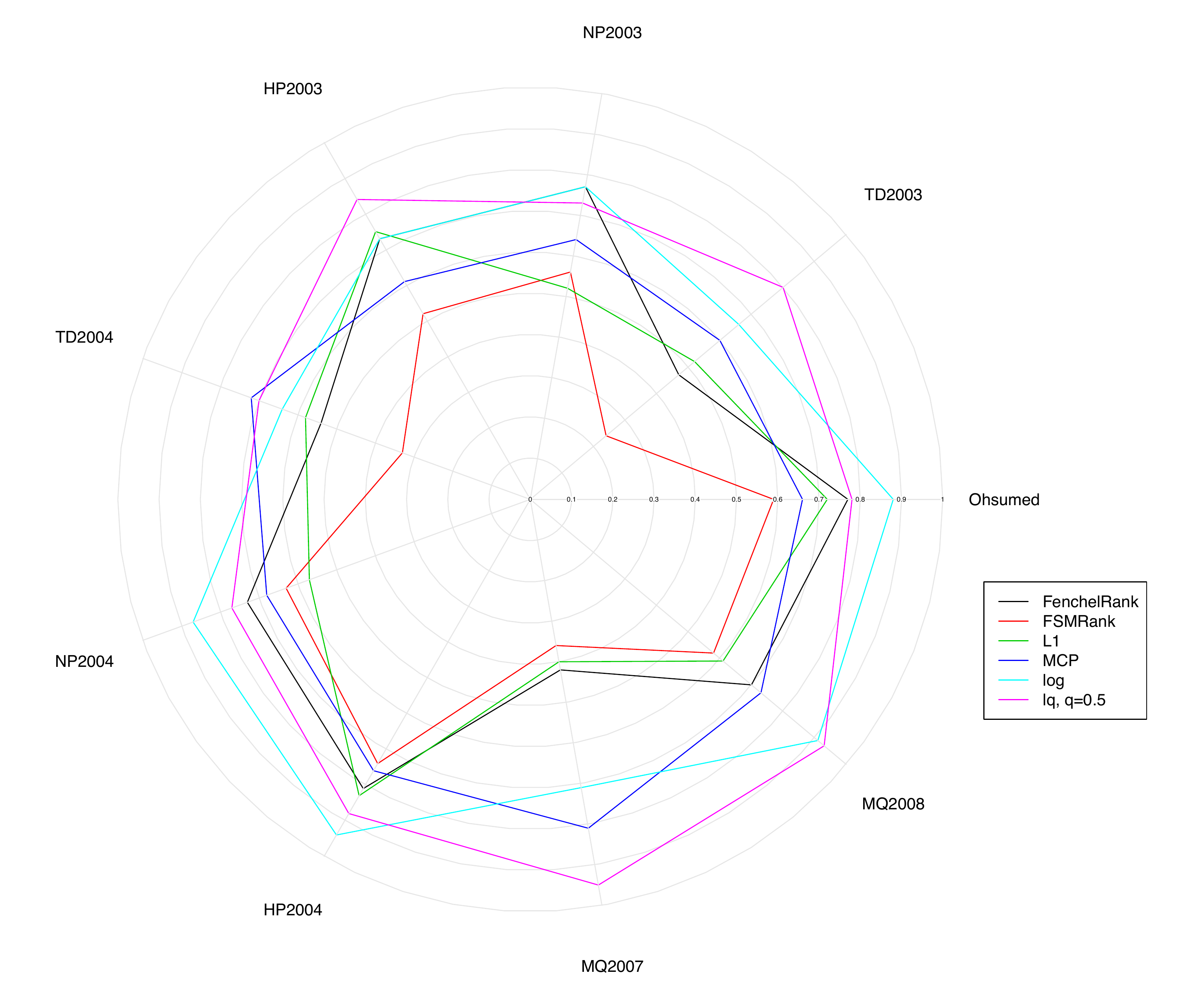}
 \caption{Ratio of removed features for each algorithm and regularization on Letor 3.0 and 4.0 corpora.}
 \label{fig:removeSparsity}
 \end{figure}

\subsection{Performance in terms of IR measures}

 In this section, we confront the prediction of our proposed frameworks to those of the two state-of-the-art algorithms FenchelRank and FSMRank. Table \ref{pvaluesMAP} (respectively table \ref{pvaluesNDCG}) indicates the algorithm that leads to the best value of MAP (respectively NDCG$@10$). Each algorithm is also compared with the best algorithm by using the unilateral one-sided Student test. If a significant decrease is observed, the percentage of degradation and the p-value are indicated. If no significant variation is observed, the two algorithms are considered as equivalent and the $\sim$ symbol is used. When considering MAP and NDCG values, one can noticed that some algorithms performs better on some datasets than on another. This is not specific to our methods and had already been observed for learning-to-rank algorithms \cite{Tieyanliu2011}.

\subsubsection{General results}All the algorithms tend to provide similar results in terms of both MAP and NDCG$@10$ on the several datasets. Nevertheless, some differences in terms of performance can be observed among the algorithms, especially on HP2004 and NP2004 datasets on which we notice the largest variations of MAP and NDCG$@10$. A more detailled study is conducted in order to determine whether this difference are significant.

\subsubsection{MAP analysis} We notice that on four datasets including Ohsumed, MQ2008, NP2004 and TD2003, all the algorithms provide similar results than the best algorithm. On the other five datasets, we observed that some convex and non-convex algorithms can lead to some degradation of the evaluation measure. We observe narrow decreases (less than 1\%) in half of the cases. Limited (between 3\% to 4\%) and higher (up to 11\%) decreases also occured. A deeper analysis follows.

FSMRank provides the higher value of MAP on the MQ2007 dataset. Our proximal approach with $\ell_1$ regularization is the only one that leads to equivalent results. We observed a very narrow decrease of the MAP when using FenchelRank (-0.8\%) and our reweighted framework with MCP and log penalties (-0.5\% and -0.7\% respectively). The use of the $\ell_p$ regularization leads a 3\% degradation of the MAP, which is still reasonable.

FenchelRank provides the best MAP results on HP2004, but all the other algorithms and regularizations provide comparable results, except the log penalty. The framework using the log regularization leads to a degradation of 11\% of the MAP, so the use of this penalty might not be a good choice in term of MAP performance on this particular dataset. When using non-convex penalties, the $\ell_p$ or MCP ones should be prefered on this dataset.

On TD2004 dataset, we observe a significant degradation of the MAP only when considering the mcp regularization. The other non-convex penalties and the $\ell_1$ regularization lead to results equivalent to the best algorithm. On NP2003 dataset, all the algorithms lead to similar results, except when using the $\ell_p$ penalty, for which a small variation is observed.

Finally, we notice that all the non-convex penalties provide as good results as our $\ell_1$ algorithm, for which the MAP is the highest. FSMRank is the only one for which a degradation is observed.

All in all, the framework we proposed leads to competitive results in terms of MAP. The $\ell_1$ algorithm is the best on one dataset and provides equivalent results on all other datasets. The MCP regularization leads to the best MAP values on two datasets and is equivalent to the best method on five datasets. The log and $\ell_p$ penalties provides results similar to the best algorithm on seven datasets.

\subsubsection{NDCG@10 analysis}When considering the NDCG$@10$, we observe that most of the algorithms provide similar results on most of the datasets. Highest NDCG$@10$ values are obtained by FenchelRank on two datasets, FSMRank on two datasets, our $\ell_1$ framework on three datasets and our framework with non-convex log penalty on two datasets. Nevertheless, all the algorithms provide similar results compared to the best algorithm on seven datasets, including Ohsumed, MQ2008, NP2004, TD2004, HP2003, NP2003 and TD2003. Thus, our framework leads to similar performance than the state-of-the-art algorithms and can provide highest NDCG$@10$ values, althought the increase is not significant.

As we notice for the MAP, FSMRank provides the best values on the MQ2007 dataset. We observe significant decreases for all the others algorithms, althought the degradations are narrow (less than 1\% in most cases). On the HP2004 datasets, our framework performs as well as the best algorithm, except for the log penalty. In this last case, we observe a 4\% decrease fo the NDCG$@10$.

Experiments thus show that our convex and non-convex frameworks provide similar results than convex state-of-the-art algorithms in most cases. They can lead to higher MAP or NDCG values, althought the increase is not statistically significant.

\subsection{Discussions}

In previous sections, we analyze independently the ability of our framework to select only a few number of features and their performance prediction. We showed that non-convex penalties are competitive to reduce the number of features used by the learned models. We also pointed out that non-convex penalties leads to similar results than the best algorithms on most datasets.

Figure \ref{fig:IRmeasures} plots the MAP values against the sparsity ratio for three representative datasets. For each dataset, the average values of MAP among all the algorithms is represented by a dotted line. We restrain the number of datasets for readability reasons. Figure \ref{fig:IRmeasures} show the use of non-convex regularizations, especially the $\ell_p$ and log penalties, are highly competitive feature selection methods, both in terms of sparsity and prediction quality. Indeed, they achieve MAP and NDCG$@10$ performances that are similar to state-of-the-art convex algorithms, while selected half as many features in average on the datasets. On most datasets, the log and $\ell_p$  penalties are the methods that select the smaller number of features, while the MAP remains stable. They can select up to six times fewer features than the other convex algorithms, without any significant degradation of evaluation measures.

In the few cases where a significant decrease is observed, the degradation is usually narrow. On MQ2007 dataset, the MAP and NDCG$@10$ degradation observed when using the log penalty is less than 1\%. It is similar to those obtained with convex algorithms, whereas the log penalty selects half as many features as the convex algorithms. On the same dataset, we observe a 3\% decrease of the MAP and a 5.6\% decrease of the NDCG$@10$ when using the $\ell_p$ penalty, but the algorithm select up to twelve times less features than the convex approaches, and up to four times fewer features than the other non-convex algorithms.

On HP2004, we observe a degradation of 11\% of the MAP and of 4\% of the NDCG when using the log penalty, but this method presents the better sparsity ratio, and selects a quarter as many features as state-of-the-art methods. On an other hand, the $\ell_p$ penalty provides as good results as the best method while selecting 37\% less features than the best algorithm. On NP2003, we observe a degradation only for the $\ell_p$ penalty, whereas the log penalty provides similar results than convex methods and uses half as many features as the best algorithm FSMRank and the same number of features than FenchelRank. On this very particular dataset, non-convex methods do not perform as well as on the others, which may be due to the specificity of this dataset.

Moreover, we do not tune the specific parameter of non-convex regularizations, but set them to default values that are usually used by the community. Results may be improved by an appropriate tuning of these parameters.

As a conclusion, the framework we proposed is able to provide similar results in terms of quality prediction compared to state-of-the-art approaches, while selecting half as many features. They are then competitive methods for feature selection in learning to rank.
 \begin{table}
  \caption{Comparison of MAP between the best method on each dataset and others algorithms. Best MAP is in bold. The $\sim$ symbol indicates equivalence between two methods. Percentage of decrease is presented when statistically significant under the 5\% threshold (p-values in italics)}
  \label{pvaluesMAP}
  \begin{center}
   \begin{tabular}{c |c c c c c c}
    \hline
    \textbf{Dataset}&\textbf{Fenchel}&\textbf{FSM}&\textbf{$\ell_1$}&\textbf{MCP}&\textbf{Log}&\textbf{$\ell_{0.5}$}\\
    \hline\hline
\textbf{Ohsumed}&$\sim$&$\sim$&$\sim$&\textbf{0,4513}&$\sim$&$\sim$\\
& & & & & & \\
\textbf{MQ2008}&\textbf{0,4804}&$\sim$&$\sim$&$\sim$&$\sim$&$\sim$\\
& & & & & & \\
\textbf{MQ2007}&-0.8\% &\textbf{0.4672}&$\sim$&-0.5\%&-0.7\%&-3\% \\
&\textit{0.02} & & & \textit{0.03} &\textit{$<$0.001}&\textit{0.02}\\
\textbf{HP2004}&\textbf{0.7447}&$\sim$&$\sim$& $\sim$ &-11\%&$\sim$\\
& & & & &  \textit{0.01}& \\
\textbf{NP2004}&\textbf{0.6946}&$\sim$&$\sim$&$\sim$&$\sim$&$\sim$\\
& & & & & & \\
\textbf{TD2004}&$\sim$&\textbf{0.2327}&$\sim$&-4.7\% &$\sim$&$\sim$\\
& & & & \textit{0.039}& & \\
\textbf{HP2003}&$\sim$&-4\% &\textbf{0.7638}&$\sim$&$\sim$&$\sim$\\
& & \textit{0.008}& & & & \\
\textbf{NP2003}&$\sim$&\textbf{0.6823}&$\sim$&$\sim$&$\sim$&-4\% \\
& & & & & & \textit{0.042} \\
\textbf{TD2003}&$\sim$&$\sim$&$\sim$&\textbf{0.2670}&$\sim$&$\sim$\\
& & & & & & \\
    \hline
   \end{tabular}

  \end{center}

 \end{table}

 \begin{table}
   \caption{Comparison of NDCG$@10$ between the best method on each dataset and others algorithms. Best MAP is in bold. The $\sim$ symbol indicates equivalence between two methods. Percentage of decrease is presented when statistically significant under the 5\% threshold (p-values in italics)}
  \label{pvaluesNDCG}
  \begin{center}
   \begin{tabular}{c | c c c c c c}
    \hline
    \textbf{Dataset}&\textbf{Fenchel}&\textbf{FSM}&\textbf{$\ell_1$}&\textbf{MCP}&\textbf{Log}&\textbf{$\ell_{0.5}$}\\
    \hline\hline
    \textbf{Ohsumed}&$\sim$&$\sim$&$\sim$&$\sim$& \textbf{0.4591}&$\sim$\\
    & & & & & &\\
     \textbf{MQ2008}&$\sim$& \textbf{0.2323}&$\sim$&$\sim$&$\sim$&$\sim$\\
    & & & & & &\\
    \textbf{MQ2007}&-1\%& \textbf{0.4445}&-0.7\%&-0.9\%&-0.5\%&-5.6\%\\
    & \textit{0.043}& &\textit{0.042} &\textit{0.013}&\textit{0.0045}&\textit{0.0016}\\
     \textbf{HP2004}& \textbf{0.8274}&$\sim$&$\sim$&$\sim$&-4\%&$\sim$\\
    & & & & &\textit{0.044}&\\
     \textbf{NP2004}&$\sim$&$\sim$& \textbf{0.8148}&$\sim$&$\sim$&$\sim$\\
    & & & & & &\\
     \textbf{TD2004}&$\sim$&$\sim$&$\sim$&$\sim$& \textbf{0.3153}&$\sim$\\
    & & & & & &\\
     \textbf{HP2003}& \textbf{0.8283}&$\sim$&$\sim$&$\sim$&$\sim$&$\sim$\\
    & & & & & &\\
     \textbf{NP2003}&$\sim$&$\sim$& \textbf{0.7839}&$\sim$&$\sim$&$\sim$\\
    & & & & & &\\
     \textbf{TD2003}&$\sim$&$\sim$& \textbf{0.3538}&$\sim$&$\sim$&$\sim$\\
    & & & & & &\\
   \hline
   \end{tabular}
  \end{center}
 \end{table}

 \begin{figure*}[!t]
\begin{center}
 \includegraphics[width=0.6\textwidth]{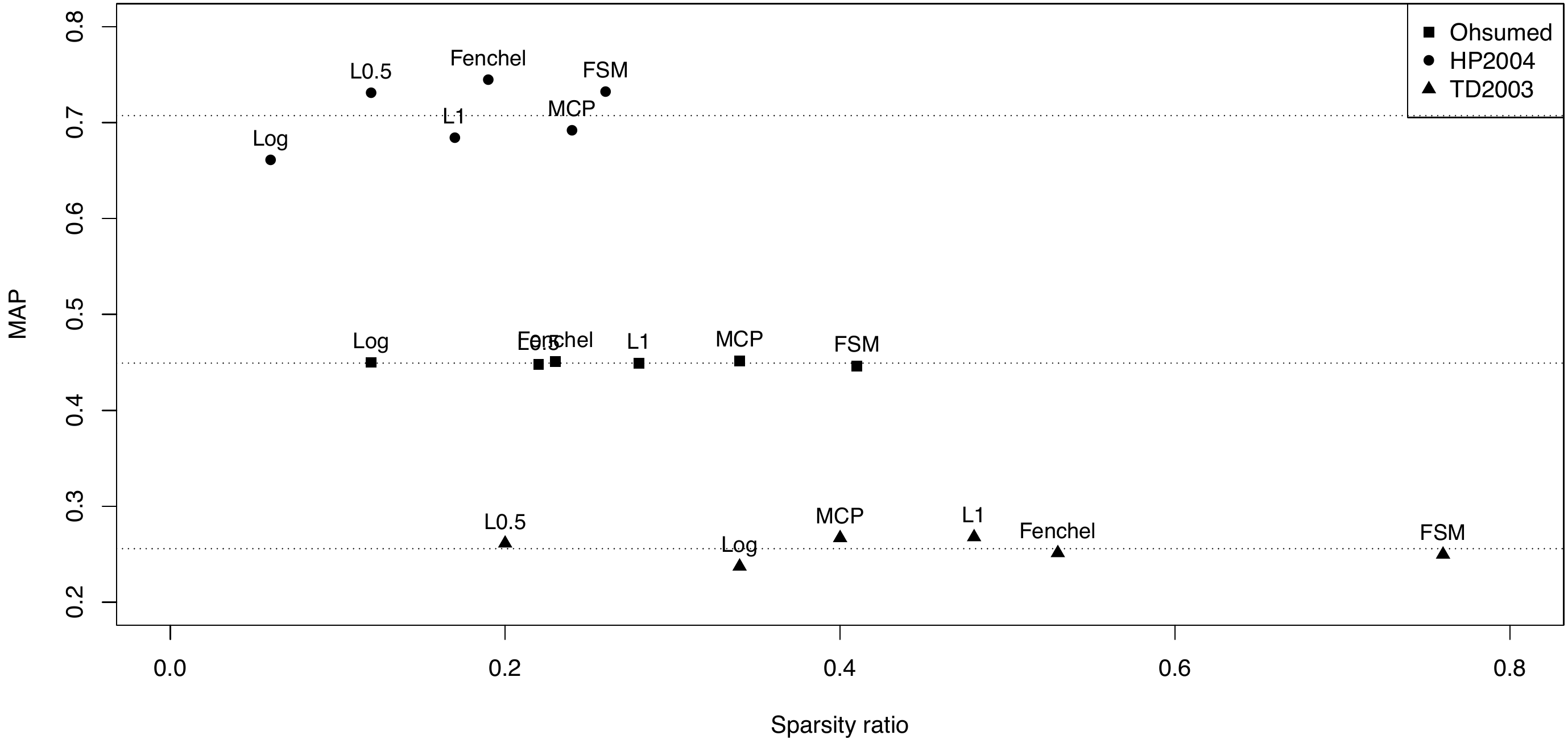}
\end{center}
 \caption{MAP \textit{sparsity ratio} for three representative datasets. Dotted lines represented average MAP obtained with the different algorithms.}
 \label{fig:IRmeasures}
 \end{figure*}
 }

\section{Conclusion and Perspectives}

\remi{In this work, we presented a general framework for feature selection in learning to rank, by using SVM with sparse regularizations. We first proposed an accelerated proximal algorithm to solve the convex $\ell_1$ regularized problem. This algorithm has the same theoretical convergence rate than the state-of-the-art FenchelRank and FSMrank algorithm. We showed that a reweighted $\ell_1$ scheme can be used in order to solve non convex problems. This scheme is implemented into a second algorithm that solved problems with MCP, log and $\ell_p$, $p<1$ penalties. To the best of our knowledge, it is the first work that propose to consider non-convex penalties for feature selection in learning to rank}  We conducted experiments on two major benchmarks in learning to rank that include nine different datasets on which we evaluate the performance in terms of MAP and NDCG$@10$. \remi{We also evaluate the ability of our framework to induce sparsity into models.}

\remi{We pointed out that the non-convex penalties lead to similar
  prediction quality, whatever the evaluation measure is, while using
  only half as many features as convex methods. Our framework is then
  a novel, competitive and effective embedded method for feature
  selection in learning to rank. Its originality is to consider
  non-convex regularizations in order to induce more sparsity into
  models without degradation of the prediction quality. Moreover, we
  will provide publicly available software for the two proposed
  algorithms in order to promote reproducible research.}

This work and the contributions of Sun \MakeLowercase{\textit{et al.}}  \cite{Sun:2009:RSR:1571941.1571987}, Lai \MakeLowercase{\textit{et al.}} \cite{10.1109/TC.2012.62}  \cite{FSMRank} show the effectiveness of embedded methods in the field of feature selection for learning to rank. More specifically, the use of sparse regularized SVM seems to be a promising way to handle the issue of feature selection and dimensionality reduction in learning to rank.  To the best of our knowledge, our work is the first that propose a feature selection framework for learning to rank that is not restricted to the use of $\ell_1$-regularization. \remi{A wide range of issues still need to be explored. In future works, we plan to evaluate the impact of tuning the non-convex regularizations parameters on both sparsity and prediction quality.  A large study of the computational times of the sparse leaning to rank algorithms could be conducted.} Finally, as feature selection can be used in order to learn ranking function specific to subset of queries, one of the most promising direction of work is the field of multitask learning. We plan to investigate the potential of a sparse regularized SVM algorithm using a Fast Iterative Shrinkage Thresholding framework, to be confronted to existing multitask algorithm \cite{Bai:2009:MLL:1645953.1646169}\cite{Chapelle:2010:MLB:1835804.1835953}\cite{Chang2012173}.

\section*{Acknowledgment}
The authors would like to thank CALMIP (grant 2012-32), the FREMIT federation and the R\'egion Midi-Pyr\'en\'ees for their support.
\hfill

\bibliographystyle{IEEEtran}

\end{document}